# Diagrams-to-Dynamics (D2D): Exploring Causal Loop Diagram Leverage Points under Uncertainty

Jeroen F. Uleman[1,*], Loes Crielaard[2],
Leonie K. Elsenburg[1,2,3], Guido A. Veldhuis[4,5],
Naja Hulvej Rod[1], Rick Quax[6], Vítor V. Vasconcelos[3,6,7]

[1] Copenhagen Health Complexity Center, University of Copenhagen, Copenhagen, Denmark.
[2] Department of Public and Occupational Health, Amsterdam UMC University of Amsterdam, Amsterdam, The Netherlands.
[3] Center for Urban Mental Health, University of Amsterdam, Amsterdam, The Netherlands.
[4] TNO - The Netherlands Organization for Applied Scientific Research, The Hague, the Netherlands
[5] Institute for Management Research, Radboud University, Nijmegen, the Netherlands
[6] Computational Science Lab, Informatics Institute, University of Amsterdam, Amsterdam, The Netherlands.
[7] POLDER center, Institute for Advanced Study, University of Amsterdam, The Netherlands.

*Corresponding author. jeroen.uleman@sund.ku.dk

**Abstract**

Background: Causal loop diagrams (CLDs) are widely used in health and environmental research to represent hypothesized causal structures underlying complex problems. However, as qualitative and static representations, CLDs are limited in their ability to support dynamic analysis and inform intervention strategies. We propose Diagrams-to-Dynamics (D2D), a method for converting CLDs into exploratory system dynamics models in the absence of empirical data. With minimal user input—following a protocol to label variables as stocks, flows or auxiliaries, and constants—D2D utilizes the structural information already encoded in CLDs, namely the existence and polarity of causal connections, to simulate hypothetical interventions and explore potentially influential places to intervene, known as 'leverage points,' under uncertainty.

Results: D2D helps distinguish between high- and low-ranked leverage points. We compare D2D to a calibrated system dynamics model constructed from the same CLD and variable labels. D2D showed greater consistency with the calibrated model than did static network centrality analysis, while also providing uncertainty estimates and guidance for future data collection.

Conclusions: The D2D method is implemented in an open-source Python package and a web-based application to support further testing and to lower the barrier to dynamic modeling for researchers working with CLDs. Future studies could help establish the approach's utility across a broad range of cases and domains.

**Background**

Complex problems arise from interacting components that form feedback loops operating across multiple scales [1, 2]. Addressing them requires moving beyond reductionist approaches and capturing the dynamic interplay between biological, psychological, and social factors [3–5]. Accordingly, complexity science has gained traction in fields like public health [3–7], psychology [8–12], and environmental sciences [13–15], offering novel ways to analyze these interconnected systems.

As a key tool in complexity science, causal loop diagrams (CLDs) are increasingly used to explore and communicate the feedback mechanisms underlying complex problems. In health sciences, areas of application have included chronic stress [16, 17], workplace well-being [18], depression [19, 20], and concussion [21]. CLDs map causal connections between key system variables to gain insight into such problems and identify 'leverage points,' i.e., intervention targets hypothesized to significantly influence system behavior [22]. The analysis of leverage points often relies on visual inspection [23] or structured qualitative approaches like Meadows' 12 Places to Intervene in a System [22], the Intervention Level Framework [24], or the Action Scales Model [25]. CLDs have also been shown to improve systems thinking and the use of information in problem analysis [26]. Yet, CLDs are inherently qualitative and static, limiting their utility for formally testing assumptions or simulating system behavior [27]. In contrast, converting CLDs into computational system dynamics models (SDMs) enables dynamic simulations over time. It allows researchers to explore interventional 'what-if' scenarios, offering a more reliable means of identifying leverage points [9].

Despite its benefits, this CLD-to-SDM conversion is challenging because it requires both computational expertise and domain-specific knowledge to formulate meaningful equations —a combination rarely found in a single individual. Moreover, the limited availability of appropriate data often hinders the development of computational models that can be fully calibrated against empirical data [9]. Together, these barriers create a methodological gap in the quantitative analysis of CLDs and the identification of leverage points, thereby constraining complexity science's contribution to addressing complex health problems.

As alternatives to full-scale computational modeling, several 'semi-quantitative' approaches have been developed to extract insights from CLDs. Fuzzy cognitive mapping, for example, assigns fuzzified strengths to connections to enable exploration of structural patterns and policy effects [28], and MARVEL incorporates strengths and delays of connections to simulate the propagation of influence from an intervention variable [23, 29]. More recently, network centrality analysis has gained popularity in health research due to its simplicity (e.g., [30–33]), requiring no additional information beyond the CLD itself. While pragmatic, centrality metrics are static, fail to capture basic CLD properties, such as connection polarity, and may lead to poor identification of leverage points [34]. Still, their rapid adoption reveals an unmet need for accessible yet more structurally faithful tools for quantitative CLD analysis.

This paper introduces Diagrams-to-Dynamics (D2D), a novel method to serve as the first step in quantitatively analyzing CLDs. D2D facilitates the transformation of a CLD into an exploratory SDM, enabling preliminary analysis even when quantitative data are unavailable and informing future data collection [27, 35]. While SDMs are ideally built using as much data as possible, the process is time-consuming and requires substantial expertise. D2D represents an intermediate step: it offers richer analysis than static approaches by simulating a simplified dynamical process, yet it is not as accurate as an empirically calibrated SDM. The rationale is that CLDs already capture valuable structural information, including, for example, whether connection polarities are positive or negative. D2D uses this information—alongside additional user input specifying which variables represent stocks, auxiliaries, and constants—to formulate the SDM. It then simulates system behavior to explore leverage points, while accounting for uncertainty due to the absence of data and guiding future data collection. By offering a systematic way to move from conceptual diagrams to exploratory computational simulations, D2D can thus help make quantitative CLD analysis more accessible to a wider audience.

## Implementation

### *2.1 Input: Causal loop diagram*

The D2D approach is applied to a CLD, in which variables are linked by directed arrows indicating hypothesized causal relationships [36]. Each connection represents a causal hypothesis and has a polarity: "+" for effects in the same direction (and additive impacts) and "-" for opposing (and subtractive) impacts [37]. Importantly, these connections can form feedback loops, showing how downstream changes influence upstream variables.

Since D2D relies heavily on the quality of the CLD's specified structure to explore system interventions, it is best applied to explanatory CLDs ("models of X") as opposed to CLDs "relevant to debate about X," where the latter tend to focus on collaboratively rethinking systems rather than precisely representing them [38]. Variables in explanatory CLDs should be potentially measurable (e.g., 'amount of green space' rather than 'environment') and non-overlapping (e.g., 'amount of green space' and 'number of trees' overlap). Furthermore, an intermediary mechanism that is implied to underlie a specific connection should not conceptually overlap with a variable either; otherwise, the connection should have been routed through that variable (e.g., the connection between 'total daily energy intake' and 'snoring' might be mediated by another CLD variable 'body weight') [9, 39]. For more on best practices in explanatory CLD development, see, e.g., [9, 36, 39–41].

Example:

To illustrate and evaluate the approach, we apply D2D to an existing explanatory CLD from Uleman et al. *[42]* on Alzheimer's disease (Figure 1A), which was previously converted into an SDM calibrated against two longitudinal datasets (Figure 1B) [42]. The model comprises 31 variables and 148 connections, resulting in numerous feedback loops of varying lengths and compositions. While this example focuses on individual-level biopsychosocial variables, D2D is also applicable at the population level and to explanatory CLDs in other domains.

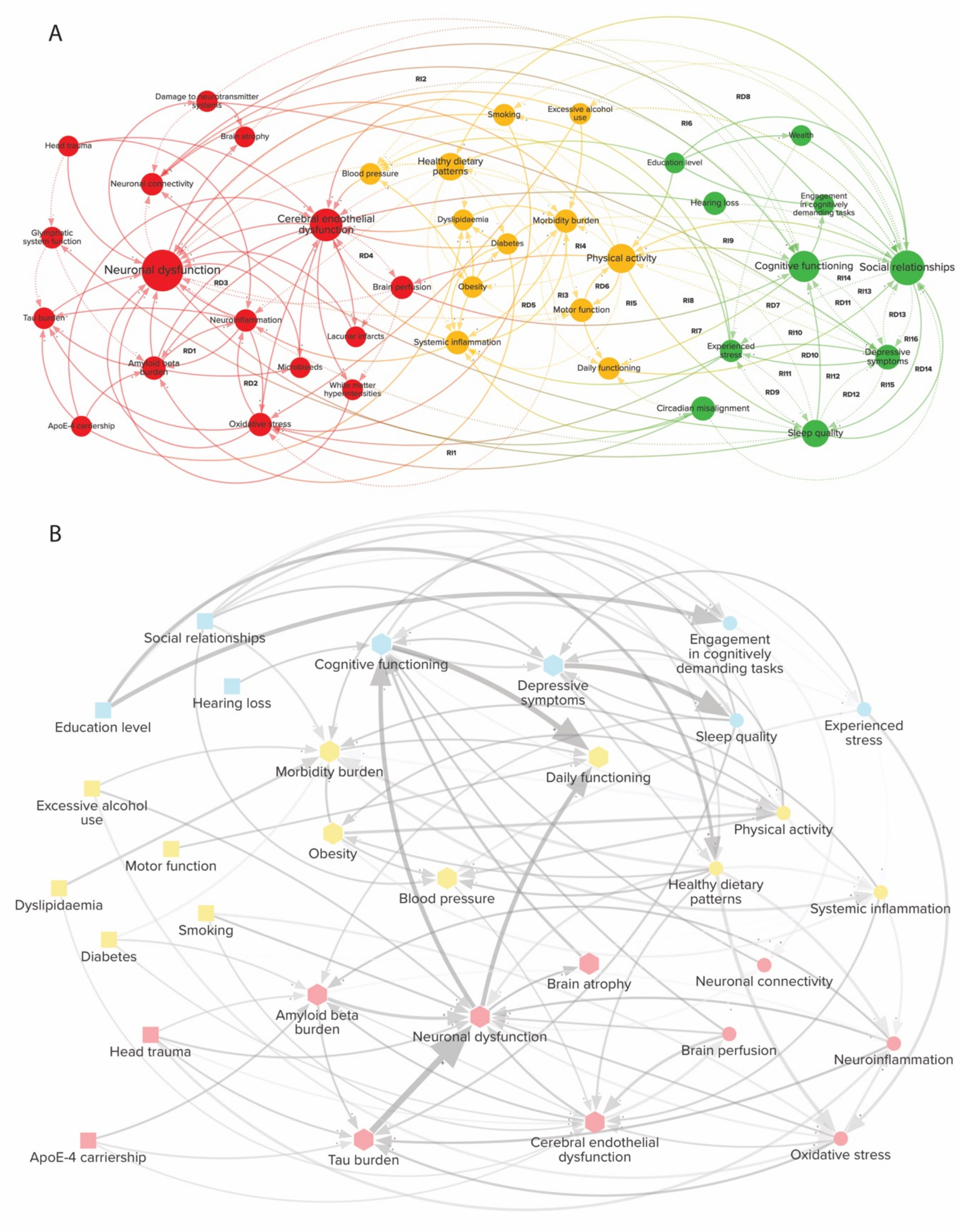

***Figure 1*** *A) The causal loop diagram on Alzheimer's disease by Uleman et al. [31]. B) A visual representation of the calibrated system dynamics model based on the causal loop diagram [42]. Constants (on the left) are indicated as squares, stocks (in the middle) are shown as hexagons, and auxiliaries (on the right) are displayed as circles. The thickness of the arrows corresponds to the strength of the connections, whereas the greyscale represents the uncertainty (darker means less*

*uncertainty). An interactive version of the diagram is available at jerul.kumu.io/simulating-the-multicausality-of-alzheimers-disease-with-system-dynamics.*

### *2.2 CLD-to-SDM conversion in D2D*

Given a CLD, a wide range of possible SDMs could be constructed depending on the types of equations used for each variable, the assumed functional forms of connections, and the values assigned to parameters within these equations. To meaningfully distinguish between potential leverage points, D2D addresses each of these sources of uncertainty by labeling CLD variables (Section 2.2.1), incorporating assumptions about the functional forms of causal relationships (Section 2.2.2), and selecting plausible ranges for parameter values (Section 2.2.3). Even in the absence of empirical data, making these assumptions enables the construction of an exploratory SDM that can support an initial analysis of potential leverage points and help guide future data collection and further model development.

#### *2.2.1 CLD variable labeling*

Labeling refers to classifying the variables into three types (constants, stocks, and auxiliaries) that determine their mathematical representation. Stocks use differential equations to model the gradual accumulation or depletion of information, material, and other states over time. Auxiliaries and flows[1] adjust instantly to inputs using algebraic equations, and constants remain fixed. As D2D treats flows and auxiliaries equivalently, we refer to both simply as "auxiliaries" hereafter.

We suggest using timescale separation[2] to perform labeling [9]. This approach considers the period over which the researcher intends to simulate the behavior of the system (the *timeframe*, e.g., disease progression over a year) and the shortest time interval over which the researcher wants to capture change in the system (the *base time unit*, e.g., monthly symptom fluctuations). Accordingly, the *timeframe* is longer than the *base time unit* to ensure meaningful dynamics. When defining these time intervals, a key consideration is the primary outcome variable(s) of interest (VOI) used to assess the effects of simulated interventions (e.g., disease outcomes).

---

[1] Traditionally, flows influence stocks, while auxiliaries impact flows. In D2D, however, we do not distinguish between flows and auxiliaries, as both are represented by algebraic equations that determine their values based on other variables at the same time point. As this distinction does not affect the model's equations or behavior, it is omitted to streamline the modeling process.

[2] For a more classical treatment of timescale separation, see, e.g., Strogatz [43] and Holling [44]

Specifically, one can consider how long (*timeframe*) and how frequently (*base time unit*) to monitor the VOI.

To label the CLD variables, the user must hypothetically decide how quickly each variable is expected to show relevant change following a sudden intervention in its causes. These timescales define, from slowest to fastest response times: constants (timescale longer than the *timeframe),* stocks (timescale similar to or longer than the *base time unit* but shorter than the *timeframe*), and auxiliaries (timescale significantly shorter than the *base time unit).* As a general rule, variables lacking ingoing connections (and thus not influenced by other variables in the model) should be labeled as constants.

When there is uncertainty in labeling, we recommend testing alternative plausible classifications to assess sensitivity. For instance, if it is unclear whether a variable should be treated as a stock, re-running simulations with it labeled as an auxiliary can help determine how much this decision influences results. If the model's behavior changes substantially, either the conclusions should be tempered accordingly, or more effort should be devoted to determining the correct label in the given context.

In addition to timescale separation, alternative labeling strategies can be used. A commonly used approach is the Snapshot Test, which classifies a variable as a stock if it would retain its value when time is hypothetically paused [45]. In practice, the Snapshot Test and timescale separation are likely to converge, as variables that are assumed to retain their values when time stops tend to be those that change on longer timescales. When discrepancies arise, we recommend testing both classifications as a sensitivity check.

Once the variables are labeled, it is important to ensure that every feedback loop contains at least one stock. Since auxiliaries are instantly determined by their causes, a loop composed entirely of auxiliaries creates a circular dependency, requiring a variable's current value to determine itself. When a feedback loop consists solely of auxiliaries, the most appropriate auxiliary—typically the one with a timescale closest to the *base time unit*—should be reclassified as a stock to maintain logical consistency in the model. When no such variable exists, the structure of the CLD may need to be reconsidered, for example, by removing one of the connections in the feedback loop, which would effectively remove the loop. If the user does not specify a label, D2D infers it based on the variable's structural position in the CLD:

variables without ingoing connections are treated as constants, whereas variables with ingoing connections are treated as stocks.

<u>Example:</u>

In the Alzheimer's disease example, *Cognitive functioning* is the VOI [42]. A 5-year *timeframe* is used, with a *base time unit* of 3 months (one quarter-year). Stocks included *Cognitive functioning, Brain atrophy, Neuronal dysfunction, Cerebral endothelial dysfunction, Amyloid beta burden, Daily functioning, Morbidity burden, Depressive symptoms, Obesity, Blood pressure, Tau burden,* with assumed timescales similar to—or longer than—3 months (but not longer than 5 years, as then they would have been constants). Auxiliaries were *Engagement in cognitively demanding tasks, Healthy dietary patterns, Physical activity, Sleep quality, Experienced stress, Systemic inflammation, Brain perfusion, Oxidative stress, Neuroinflammation, Neuronal connectivity*, with assumed timescales significantly shorter than 3 months. Finally, *Head trauma, ApoE-4 carriership,* and *Education level* were treated as constants due to their lack of inputs in the CLD (Figure 1A). In contrast, *Diabetes, Dyslipidaemia, Social relationships, Hearing loss, Smoking, Excessive alcohol use,* and *Motor function* were treated as constants despite having inputs in the CLD because their assumed timescales exceeded the 5-year simulation time, and the variables showed minimal change over those 5 years in the data *[42]*. These assumptions are summarized in the timescale separation map [9, 42] in Figure 2.

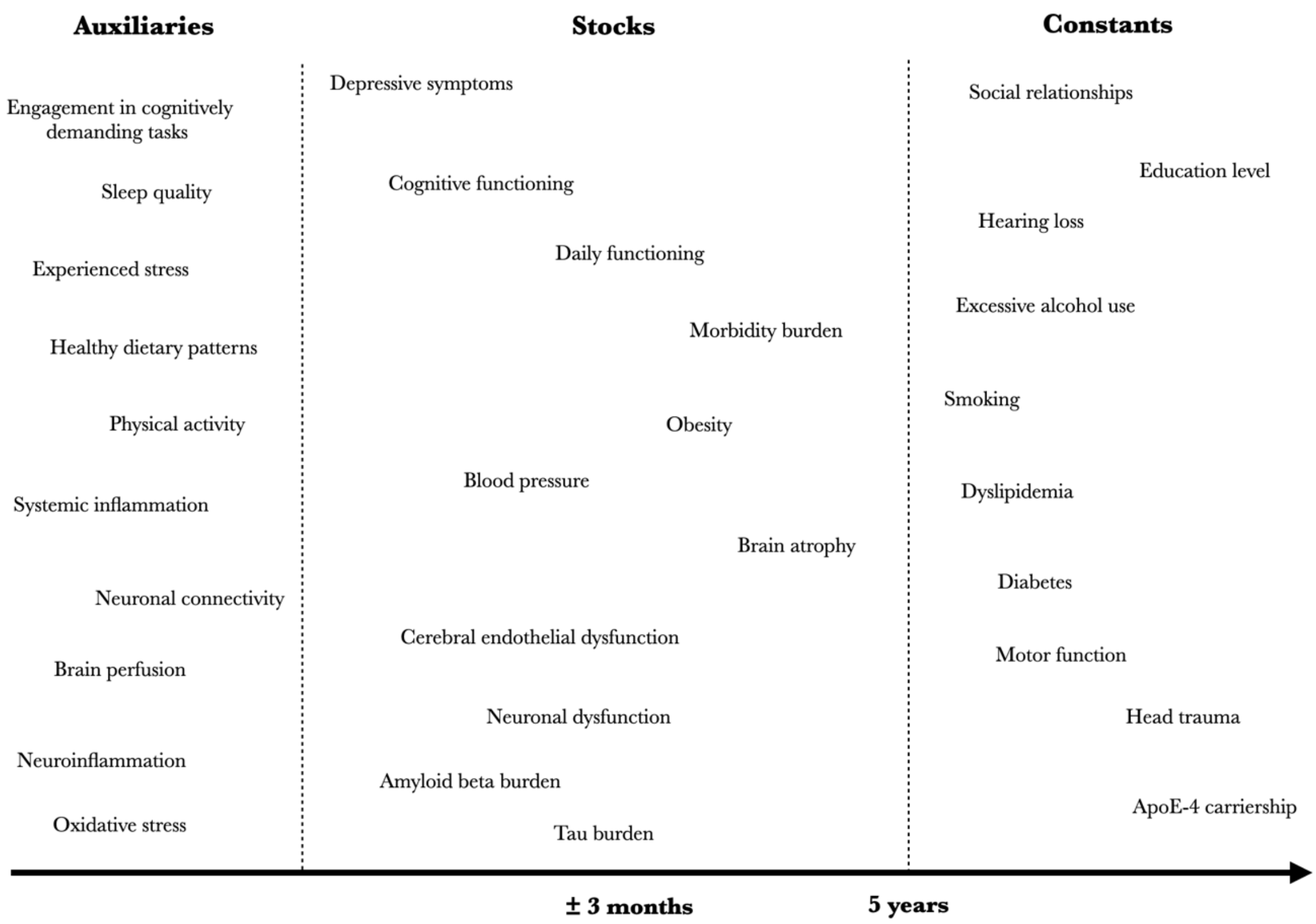

***Figure 2*** *Timescale separation map from our Alzheimer's disease case example using a base time unit of quarter-years (3 months) and a timeframe of 5 years*

*2.2.2 Defining and solving the equations of the system dynamics model*

After the CLD variables have been labeled, D2D reduces the uncertainty regarding the many possible SDM equations by assuming all CLD connections represent pairwise interactions. These connections are then modeled as additive linear terms—reflecting standard CLD practice, where functional forms are rarely specified [9]. This approach simplifies model construction and requires no additional modeling expertise or domain knowledge. Despite their simplicity, linear models can still produce nonlinear curves over time, such as exponential growth or decay, through feedback loops. Although D2D also supports adding interaction terms, we do not discuss them here and focus on the properties of the linear model. An important consequence of using linear equations is that the model can describe only a single equilibrium.

Under this linearity assumption, D2D converts the CLD into a coupled system of algebraic equations (for auxiliaries) and ordinary differential equations (for stocks). This system can be rewritten (see Additional file 1) as a system of linear ordinary differential equations (Equation 1),

$$\frac{d\boldsymbol{x}}{dt} = \boldsymbol{A}\boldsymbol{x} + \boldsymbol{b} \quad (1)$$

where the parameter matrix **A** and intercept vector **b** incorporate the stocks' and auxiliaries' coefficients and constants. Although this system can be solved analytically, the solution cannot be evaluated when $\boldsymbol{A}$ is singular. For consistency, D2D therefore uses a numerical solver.

*2.2.3 Uncertain model parameters*

The next and equally important source of uncertainty stems from the unknown strength of connections, i.e., the model parameters. To constrain and quantify this uncertainty, D2D samples parameter values from uniform distributions informed by the connection polarities in the CLD: negative connections from Uniform[$-\theta_{max}$, 0], positive connections from Uniform[0, $\theta_{max}$], and ambiguous or unspecified connections from Uniform[$-\theta_{max}$, $\theta_{max}$], where $\theta_{max}$ is the maximum strength of a connection (discussed below). This approach incorporates the polarity information but retains as much uncertainty as possible about the precise value.

To deal with the fact that the units and typical value ranges of all variables are unknown, the simulations are interpreted in standardized units (Z-scores). For auxiliary equations, a parameter value of 0.1 means that a one standard deviation (SD) change in a variable X results in a direct 0.1 SD change in auxiliary Y. For stock equations, this same parameter value implies a 0.1 SD change per *base time unit* in stock Y. Consequently, these parameters have to be treated differently, as the stock equations consider change per time unit. In contrast, the auxiliary equations consider direct change to a new stable value. For auxiliaries, the $\theta_{max}$ is independent of the model's temporal context and can be set by considering what standardized regression coefficients are reasonably expected. A good reference value could be around 0.3-0.5, depending on findings from causal inference studies in one's scientific field.

For stock equations, on the other hand, the upper bound $\theta_{max}$ depends on the model's temporal context: its *base time unit* and *timeframe*. For example, a monthly model over five years (60 steps) requires a smaller $\theta_{max}$ than a yearly model over the same period (5 steps) to show the same total change in the VOI, since each update captures a smaller fraction of the overall change. Using the same $\theta_{max}$ when changing a model's *base time unit* from years to months could cause the model to reach unrealistic values, including infinity, and would significantly increase the uncertainty in simulated intervention outcomes. On the other hand, $\theta_{max}$ should not be set too low, as this could overstate confidence and lead to misleading conclusions. We suggest the following heuristic: $\theta_{max}$ ≈ (expected total change in VOI within the timeframe) ÷ (number of base time units per timeframe). For a VOI expected to change monotonically (i.e., only in- or decreases) by ±2 SDs over 5 years (60 months), $\theta_{max}$ = 2 ÷ 60 = 0.033 per month. However, if the VOI fluctuates periodically—e.g., rises and falls 3 times to ±1 SD—its total movement could be ~6 SDs, giving $\theta_{max}$ = 6 ÷ 60 = 0.1. If the parameter settings result in unreasonably high or low values, e.g., <0.5 or >10 SD change in the VOI over the simulation time, we suggest adjusting the settings[3]. Additionally, since the $\theta_{max}$ selection can impact the simulation output, we suggest experimenting with different values to assess the output's robustness to these settings. Finally, if additional information is available for specific connections (e.g., based on scientific literature), the user can also select a range of parameter values to uniformly sample from (e.g., Uniform[0.2, 0.5]), which will override $\theta_{max}$ for these connections.

Example:

In our example, we set a $\theta_{max}$ of 0.1 for stock variables, based on expected changes in *Cognitive functioning* as measured by the ADAS-cog scale [46]. Among cognitively unimpaired but at-risk older adults, ADAS-cog scores decline by about one point per year, and one standard deviation corresponds to roughly three points. This translates to an annual decline of approximately 1/3 SD, or less than 2 SDs over five years. Since five years equals 20 quarters (our model's *base time unit*), we set $\theta_{max}$ = 2 ÷ 20 = 0.1. For auxiliary variables, we chose a $\theta_{max}$ of 0.3 because using 0.5 caused most simulations to exceed 10 SDs in the VOI and 0.3 corresponds to the largest empirically estimated auxiliary parameter (i.e., typical standardized regression coefficient) from Uleman et al. [42].

### *2.3 Analysis*

#### *2.3.1 Simulation of interventional "what-if" scenarios*

[3] Because linear systems of ordinary differential equations produce exponential trajectories, the provided heuristic offers only a first approximation of $\theta_{max}$. In practice, the realized change in the VOI depends on the interaction structure and feedback loops in the system, which may amplify or dampen effects non-linearly. Therefore, after running the simulations, D2D issues a warning when the maximum absolute deviation of the VOI over time exceeds 10 SDs, or never exceeds 0.5, indicating potentially unrealistic dynamics. Users can use these diagnostics to iteratively refine $\theta_{max}$.

Based on the specified CLD and the above assumptions, D2D defines an exploratory SDM. D2D then simulates 'what-if' scenarios that explore the effect that hypothetical interventions on a set of 'intervention variables' may have on the VOI(s) at the final timepoint of the simulation, given a perturbation of one standard deviation of each intervention variable.[4]

To simulate an intervention in D2D, we perturb the intervention variable from an equilibrium baseline (i.e., zero). For stocks, this involves setting the initial value to +1 or –1; for auxiliaries, we add -1/+1 to their equations; and for constants, we increase their value by -1/+1 for the duration of the simulation. This represents a one standard deviation shift from a completely average individual—defined as having a standardized mean of zero on all variables—toward the specified direction of interest. For instance, for *Physical activity*, we expect a higher value to be beneficial, whereas for *Smoking* we expect a lower value to be beneficial.

If desired, D2D allows the user to specify variable-specific intervention strengths to reflect how readily each variable can be altered through intervention. For example, if it is considered much more feasible to substantially increase a person's *Physical activity* than to reduce their *Poor dietary patterns*, one might assign an intervention strength of +1 to *Physical activity* and -0.5 to *Poor dietary patterns*. As such differences in assumed intervention strengths can substantially influence the resulting ranking of interventions, we recommend using this feature only when there is a strong empirical or theoretical justification. When interaction terms are present, double-factor interventions can also be simulated, with a default strength of 0.5 per intervention variable, reflecting an equal split of a total one standard-deviation effort.

Each intervention is repeated N times using independently sampled parameter sets to capture model uncertainty. This implies that each connection in the model is weak in some simulations and strong in others, reflecting our uncertainty about connection strength. We suggest using at least 100 samples, which can be adjusted as needed until the results of interest stabilize, which can be assessed using the provided bootstrap confidence intervals. D2D then ranks the median effects of interventions (i.e., in the typical effect of the intervention across the range of samples), displaying uncertainty ranges to facilitate comparison.

[4] In D2D, intervention effects are evaluated relative to a no-intervention baseline simulation, in which all variables remain at their equilibrium values (i.e., zero). Thus, the reported VOI at the end of the simulation reflects the deviation from this baseline following a one–standard deviation perturbation of the intervention variable.

Example:

In our case example, we intervene on 15 modifiable risk factors within the CLD. Specifically, we decrease *Obesity*, *Dyslipidaemia*, *Blood pressure*, *Diabetes*, *Smoking*, *Depressive symptoms*, *Hearing loss, Head trauma*, and *Excessive alcohol use*, and we increase *Education level*, *Physical activity*, *Social relationships*, *Healthy dietary patterns*, *Engagement in cognitively demanding tasks,* and *Sleep quality.* We simulated interventions on each of these variables $N$ = 200 times, using random samples from the model parameters. The code for the specific analyses of our case example can be found here:
http://github.com/jerul/systemdynamics/blob/main/tutorials/Alzheimer.ipynb.

The resulting ranking in Figure 3 shows how D2D can differentiate between high- and low-ranked intervention variables. As can be seen, *Sleep quality* and *Depressive symptoms* rank the highest by their median effect on change in *Cognitive functioning*, while *Dyslipidaemia*, *Blood pressure,* and *Engagement in cognitively demanding tasks* rank the lowest. All intervention effects fall on the right-hand side of the 0 line, indicating that each had a beneficial effect and resulted in less cognitive decline than would have otherwise occurred. However, the uncertainty in these rankings is substantial, and it would be impossible to identify a definitive set of potential leverage points directly from D2D. For example, simulated interventions on *Physical activity* had greater impacts than *Sleep quality* in 21% of the samples (Table 1).

Nevertheless, one might draw careful conclusions at the extremes of the ranking, or at least provide a reduced set of variables to focus on further. For instance, based on these exploratory simulations, the structure of the CLD, combined with the variable labeling, indicates that intervening on *Sleep quality* could be more important than intervening on *Obesity* in affecting 5-year cognitive decline (100% of the samples). Although the uncertainty would have to be reduced to distinguish among other variables in the ranking, this analysis provides an early indication of the relative importances and likelihoods of variables as key leverage points.

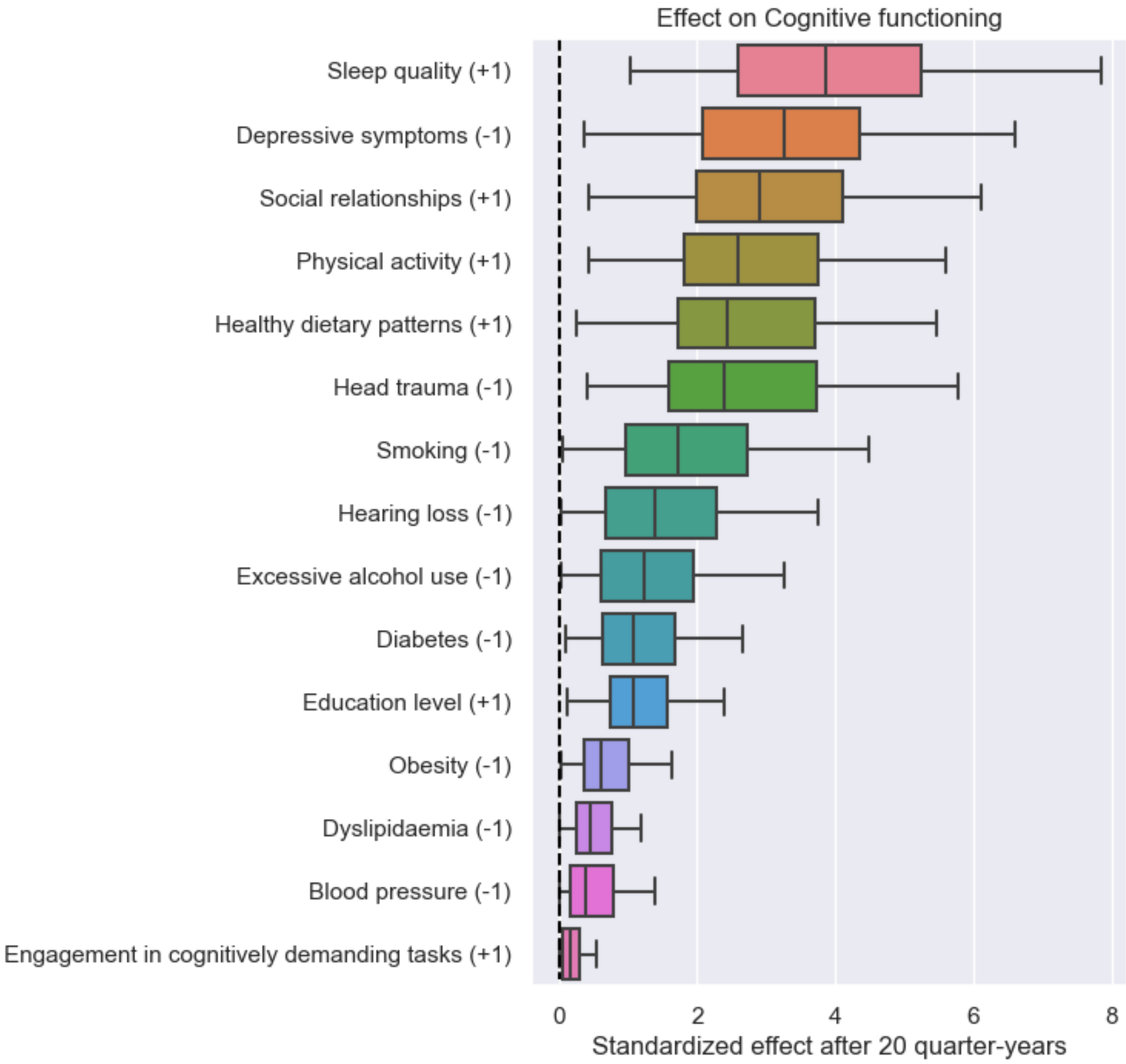

***Figure 3*** *Ranking of simulated interventions on Cognitive functioning using D2D within a timeframe of five years (20 quarter-years, which was the base time unit). The directions of the interventions, as specified by the user, are indicated on the y-axis.*

| **Intervention A** | **Intervention B** | **% A>B** | **95% CI** |
|---|---|---|---|
| Sleep quality | Depressive symptoms | 73% | [67%, 79%] |
| Sleep quality | Head trauma | 75% | [69%, 81%] |
| Sleep quality | Healthy dietary patterns | 76% | [71%, 81%] |
| Sleep quality | Social relationships | 78% | [72%, 84%] |
| Sleep quality | Physical activity | 79% | [73%, 85%] |
| Sleep quality | Smoking | 91% | [87%, 94%] |
| Sleep quality | Hearing loss | 94% | [90%, 97%] |
| Sleep quality | Excessive alcohol use | 96% | [94%, 99%] |
| Sleep quality | Diabetes | 98% | [95%, 99%] |
| Sleep quality | Education level | 100% | [100%, 100%] |
| Sleep quality | Obesity | 100% | [100%, 100%] |
| Sleep quality | Dyslipidemia | 100% | [100%, 100%] |
| Sleep quality | Blood pressure | 100% | [100%, 100%] |
| Sleep quality | Engagement in cognitively demanding tasks | 100% | [100%, 100%] |

***Table 1*** *The fraction of samples where an intervention on one variable was larger than on another with a 95% bootstrapped confidence interval (CI) based on 200 resamplings with replacement. Note: while D2D provides comparisons across all interventions, only comparisons with the top-ranked intervention variable, Sleep quality, are shown here for the sake of brevity.*

*Comparison to the calibrated SDM*

For comparison, the ranking from the calibrated SDM is shown in Figure 4 (left panel) [42]. Notable differences include *Hearing loss* ranking higher in the calibrated SDM than in D2D, while *Social relationships* and *Physical activity* rank lower, which can be explained by their relatively strong and weak impacts in the data, respectively. D2D does not make these distinctions between strong and weak impacts, as it samples uniformly from polarity-informed distributions, assuming there are no data to inform which connections are stronger or weaker. There is also significant overlap: *Sleep quality* and *Depressive symptoms* rank highest in both approaches, and factors such as *Head trauma*, *Healthy dietary patterns*, and *Smoking* are ranked as moderately influential. Note also that the uncertainty in the calibrated SDM ranking is considerably lower, as expected since it is calibrated on data. On the other hand, developing and calibrating the full SDM took about two years. This clearly shows the trade-off between effort and differentiation between interventions.

*Comparison to network centrality analysis*

Next, we compare D2D to the common practice of applying network centrality analysis, focusing first on the rankings they provide, then on conceptual alignment. Figure 4 presents the D2D results alongside the betweenness and closeness centrality scores of the CLD variables. Betweenness centrality (rightmost panel) applies only to variables with both incoming and outgoing connections, which excludes a substantial number of variables (i.e., all constants) from being assigned a non-zero value in this example. It also identifies *Physical activity* as the most influential variable. Closeness centrality, in turn, ranks *Social relationships* and *Physical activity* higher than *Sleep quality* and *Depressive symptoms*, and assigns relatively high values to variables like *Obesity* and *Education level*, which rank low in both D2D and the calibrated SDM. While closeness centrality aligns with D2D on many of the high-ranking variables, the overall pattern suggests that centrality metrics diverge more strongly from the calibrated SDM than D2D does: the Spearman's correlation between the calibrated SDM and D2D's median ranking was ρ = 0.82, compared to ρ = 0.51 for closeness and ρ = 0.11 for betweenness centrality. Importantly, centrality analysis also lacks the nuance provided by D2D's uncertainty estimates.

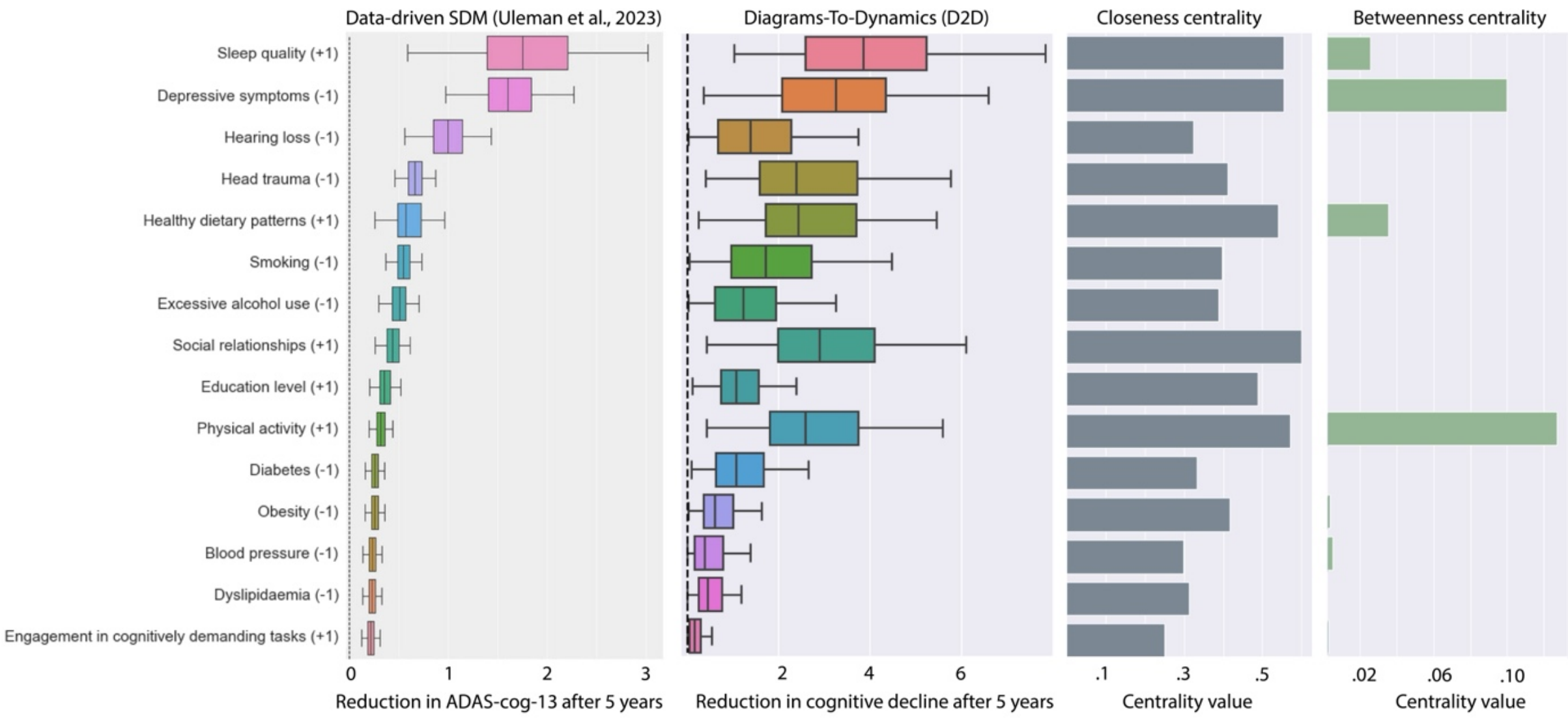

***Figure 4*** *Ranking of simulated interventions on Cognitive functioning within a timeframe of five years using the calibrated SDM (left panel) [42] and D2D (second left panel). The directions of the interventions, as specified by the user, are indicated on the y-axis. Note that the D2D results are the same as in Figure 3 but ordered by the calibrated SDM's ranking. Network centrality analysis rankings are presented for closeness centrality (second right panel) and betweenness centrality (right panel).*

Table 2 summarizes key conceptual differences between network centrality analysis and D2D, addressing critiques outlined by Crielaard et al. [34]. Centrality metrics like betweenness and closeness evaluate variable importance based on static graph properties. They may rely on undirected networks, ignore connection polarities, and assume that influence spreads along the shortest paths between variables. D2D addresses these shortcomings by simulating influence strictly in the indicated causal direction, incorporating connection polarities by sampling from polarity-constrained parameter ranges, and simulating all causal pathways rather than prioritizing only the shortest ones. D2D also enables direct exploration of interventions on system variables in a specified direction with respect to one or multiple VOI(s), making it suitable for policy design questions (under uncertainty). Although its linear formulation assumes additive effects, D2D accomodates interaction terms when users wish to examine how variables and interventions may combine non-additively—something network centrality metrics do not capture.

| **CLD considerations** (Crielaard et al. [34]) | **Centrality analysis** (Crielaard et al. [34]) | **D2D** |
|---|---|---|
| CLDs rely on arrows to describe the directions of causal effects. | Undirected variants of betweenness- and closeness centrality are sometimes used. | D2D models the effects in the CLD's indicated direction of causality. |
| CLDs rely on polarities to describe a positive or negative causal effect. | Betweenness- and closeness centrality do not take polarities into account. | D2D accounts for polarity by sampling only from the side of indicated polarity in the CLD (unless polarity is unspecified). Additionally, D2D simulates interventions in a specific direction (e.g., reducing smoking or increasing physical activity) and of a specific strength (e.g., -0.5 for one variable and +1 for another). |
| What flows through CLDs may not take the shortest path. | Betweenness- and closeness centrality assume that what flows through the network takes the shortest path. | D2D does not rely on a shortest path assumption but simulates propagation through all paths indicated by the CLD. |
| CLDs include factors from many different domains. | Betweenness- and closeness centrality assume that all factors in the CLD belong to the same domain and act at the same scale. | D2D assumes timescale separation and does not assume fixed parameter values or relative strengths of connections; instead, it samples from this uncertainty, allowing different connections to dominate in different simulations.<br><br>Consequently, variables of different domains that act at the same timescale (e.g., environmental and biological) are treated equally, as there is no empirical basis for weighting them differently. |
| CLDs are frequently used to show interactions between lower and higher domains, which may conceptually overlap. | Betweenness- and closeness centrality assume that there is no overlap between the factors in the CLD. | D2D also assumes conceptually distinct variables, as overlapping constructs can distort inferences when simulating interventions. It should therefore be exclusively applied to explanatory CLDs that represent models of a system, where node distinctiveness is crucial. |
| CLDs are developed to inform complex interventions with interacting components. | Betweenness- and closeness centrality cannot tell us how interventions on different factors interact. | D2D simulates a range of interventions relative to an outcome variable of interest. When multiple interventions are simulated simultaneously in a linear model, their effects are additive. However, the user can also add interaction terms and, if so, intervening on two variables simultaneously may have a different effect than intervening on them in isolation. Therefore, D2D can simulate two-factor interventions when interaction terms are included. |

***Table 2*** *Identified problems with network centrality analysis applied to causal loop diagrams using betweenness and closeness centrality [34] and how these relate to the Diagrams-to-Dynamics (D2D) approach.*

*2.3.2 Sensitivity analysis to determine how to best reduce uncertainty in the 'what-if' scenarios*

Given the substantial uncertainty in the simulated interventions, the next step in the D2D analysis phase is to conduct a sensitivity analysis. Using each parameter sample, D2D assesses the impact of the model parameters on the change in the VOI, which is the same quantity used to assess the effect of the simulated interventions. If this quantity is particularly sensitive to fluctuations in specific parameters, narrowing the uncertainty in these parameters could significantly reduce its uncertainty, enabling more precise identification of the simulated interventions. Thus, targeted data collection on the relevant variables would allow us to draw stronger conclusions regarding system leverage points[5]. D2D uses the monotonic Spearman's correlation coefficient (ρ) to represent the sensitivity.

Example:

In our example, Table 3 summarizes the parameters to which changes in *Cognitive functioning* were most sensitive, aggregated across all simulated interventions. Under the current assumptions, key influences include effects to and from *Neuronal dysfunction*, *Neuronal connectivity*, *Depressive symptoms*, *Hearing loss*, *Brain atrophy*, and *Cognitive functioning* itself. Prioritizing these variables in future data collection would help reduce uncertainty in the VOI, even if some less influential parameters remain unmeasured.

However, the aggregated sensitivities in Table 3 implicitly weight all intervention scenarios equally, while, in practice, the sensitivity of the VOI to a given parameter can vary substantially across interventions. For instance, the effect of *Depressive symptoms* on *Cognitive functioning* has a mean ρ of −0.21 across interventions, but with a standard deviation of 0.24. This variability indicates that while this connection is influential for some interventions, it may be negligible for others. It suggests that, once a researcher has narrowed the set of interventions of interest (e.g., based on real-world feasibility), the sensitivity analysis should be examined at the level of individual interventions rather than in aggregate. Table 4 illustrates this for the highest-ranked intervention on *Sleep quality*. For this intervention, the most sensitive parameters are *Sleep quality → Cognitive functioning* (ρ = 0.64), *Depressive symptoms → Cognitive functioning* (ρ = -0.42), and *Sleep quality → Depressive symptoms* (ρ = -0.38). This pattern suggests that collecting data on these specific variables would be most effective for reducing uncertainty about the simulated effect of improving sleep on cognitive decline.

| **Connection** | **Overall mean ρ with 95% CI** | **Mean ρ per intervention** | **SD ρ per intervention** |
|---|---|---|---|
| Neuronal dysfunction → Cognitive functioning | -0.24 [-0.27, -0.20] | -0.37 | 0.18 |
| Depressive symptoms → Cognitive functioning | -0.13 [-0.16, -0.09] | -0.21 | 0.24 |
| Hearing loss → Cognitive functioning | -0.11 [-0.14, -0.08] | -0.15 | 0.22 |
| Neuronal connectivity → Cognitive functioning | 0.10 [0.07, 0.14] | 0.18 | 0.13 |
| Brain atrophy → Cognitive functioning | -0.09 [-0.12, -0.06] | -0.12 | 0.09 |
| Neuronal dysfunction → Neuronal connectivity | -0.09 [-0.12, -0.07] | -0.15 | 0.07 |

[5] Data can be incorporated following a Bayesian approach, see section 3.2 in the Discussion

| | | | |
|---|---|---|---|
| Sleep quality → Cognitive functioning | 0.08 [0.05, 0.12] | 0.14 | 0.18 |

***Table 3*** *Sensitivity analysis of the model parameters with respect to the change in the variable of interest (Cognitive functioning) over all simulated interventions. Assessed using Spearman's ρ correlation coefficient. For brevity purposes, only the parameters with* $|\rho| \geq 0.08$ *are shown. The global ρ is calculated across all parameter samples and interventions, whereas the mean and standard deviation (SD) of ρ per intervention are obtained by first averaging ρ over parameter samples within each intervention, and then taking the mean and SD across interventions.*

| **Connection** | **ρ with 95% CI** |
|---|---|
| Sleep quality → Cognitive functioning | 0.64 [0.55, 0.71] |
| Depressive symptoms → Cognitive functioning | -0.42 [-0.54, -0.31] |
| Sleep quality → Depressive symptoms | -0.38 [-0.48, -0.27] |
| Depressive symptoms → Sleep quality | -0.21 [-0.33, -0.07] |
| Neuronal dysfunction → Cognitive functioning | -0.20 [-0.33, -0.08] |

***Table 4*** *Sensitivity analysis of the model parameters with respect to the change in the variable of interest (Cognitive functioning) for the top-ranked simulated intervention (Sleep quality). Assessed using Spearman's ρ correlation coefficient. For brevity purposes, only the parameters with* $|\rho| \geq 0.2$ *are shown.*

*2.4 Software*

We provide an easy-to-use Python script to run D2D, implemented in the systemdynamics Python package. The method can also be accessed via an online Streamlit app. Links to both the repository and the app are available here: https://doi.org/10.17605/OSF.IO/4UTGW. A CLD can be loaded into D2D using an Excel file that follows the diagram export template of the widely used Kumu diagramming software (36). We added annotations to these templates that facilitate the CLD-to-SDM conversion process. This Kumu template is provided and explained in detail in Additional file 2.

## Discussion

While a major motivation for developing CLDs is to identify leverage points for intervention, commonly used static quantitative methods, such as network centrality analysis, can lead to misleading conclusions [34]. In this paper, we present D2D, a semi-automated approach for turning CLDs into exploratory system dynamics models when quantitative data are not (yet) available or yet to be collected. D2D transforms a CLD into an SDM, samples feasible parameter value combinations, simulates hypothetical "what-if" interventions, and assesses the sensitivity of results to parameter uncertainty. These sensitivities can inform downstream steps such as selecting opportune data sets, designing data collection, or zooming in on a subset of potential leverage points.

### *3.1 Using D2D to guide future data collection efforts*

By combining simulated interventions and parameter sensitivity analysis, D2D can help prioritize variables for future data collection. Preliminary intervention simulations may help identify which variables, when perturbed, most strongly affect the VOI(s), highlighting potential leverage points. Sensitivity analysis complements this by revealing which parameter uncertainties seem to drive the greatest variability in the simulations. Variables that are influential in intervention simulations and/or sensitive in parameter sensitivity analysis could be prioritized for future data collection, as better data on these variables would help reduce uncertainty efficiently and improve the precision of leverage point identification.

### *3.2 D2D within an iterative workflow*

In practice, D2D can be used iteratively to progressively refine the analysis as additional information becomes available. A researcher might begin with default parameter settings to obtain an initial ranking and sensitivity analysis, then consult available literature or data to constrain the parameter ranges of the most sensitive connections. D2D can be re-run with these updated bounds, yielding more precise intervention rankings. Where domain knowledge warrants it, interaction terms can be introduced to capture specific relationships, thereby relaxing the default linearity assumption. In this way, D2D supports a gradual transition from a qualitative CLD toward a calibrated SDM, with each iteration reducing uncertainty and incorporating additional knowledge about the system. Moreover, D2D can also serve as a tool for CLD refinement. When the ranking of a variable diverges from expert expectations, this may indicate that the CLD's structure does not adequately capture the variable's hypothesized influence on the outcome; for example, because a direct connection is missing or because an important pathway is mediated through too many intermediary variables. Such discrepancies can prompt a re-examination of the CLD, leading to a more carefully specified causal structure before further modeling is undertaken.

### *3.3 D2D within a Bayesian Framework*

D2D naturally aligns with a Bayesian perspective [47, 48]. The uncertainty over model parameters can be interpreted as a prior distribution that represents experts' beliefs about the connections' polarities before considering empirical data. D2D then essentially performs prior predictive modeling, simulating system outcomes from samples drawn from this prior [48]. When longitudinal data become available—for instance, panel data tracking individuals over time—this prior can be updated to generate a posterior distribution over parameter values, thereby reducing uncertainty. Future work will involve a broader Bayesian framework for developing SDMs from CLDs, and extensions of our Python package will incorporate Bayesian sampling methods such as Hamiltonian Monte Carlo [47] and Amortized Bayesian Inference [49] to facilitate this updating process. D2D can then be seen as a natural first explorative step in this framework, where a restricted range of functional forms is used and empirical data are not yet considered.

*3.4 Considerations in comparing D2D to network centrality analysis and the calibrated SDM*

Overall, D2D provided more informative insights than network centrality analysis, such as uncertainty estimates and guidance for future data collection, fit better conceptually with CLDs (see Table 2), and aligned more closely with the calibrated SDM. However, this alignment should be interpreted with caution, as both system dynamics modeling implementations share the same assumptions about the functional forms of connections and the signs of parameters, which may have influenced the similarity of their results. Both D2D and the calibrated SDM relied on linear equations, whereas fully developed SDMs often incorporate nonlinear relationships. Additionally, the calibrated SDM also restricted the parameters to the connection polarities using an informative prior, making its rankings more similar to D2D's than they might have been with a broad, uninformative prior. Despite these similarities, key differences remain, suggesting that there are other reasons that the methods arrive at similar results and that D2D can serve as an informative first approximation. D2D did not adjust for external factors such as age and gender, which were incorporated into the calibrated SDM. More importantly, the parameters in the calibrated SDM are estimated from extensive empirical data, resulting in lower uncertainty than in D2D (Figure 4). Overall, the example demonstrates D2D's potential under certain conditions, but further case studies are needed to assess its broader applicability and relevance across scientific domains.

*3.5 Relating D2D to other CLD quantification methods*

D2D is not the first method to provide quantitative insight into the complex phenomena described by CLDs; however, it differs from existing approaches. Unlike MARVEL, which enriches CLDs with strengths and delays to produce qualitative 0–1 influence trajectories [29], and unlike fuzzy cognitive mapping, which propagates signed weights through a smoothing function without a time base or delays [28], D2D converts an explanatory CLD into a linear

SDM (unless interaction terms are provided) by labeling variables by timescale. It then samples from polarity-constrained parameters (if polarity is provided) and solves the resulting ordinary differential equations analytically (or numerically when system matrix $\boldsymbol{A}$ is non-invertible or interaction terms are provided). This workflow preserves the accessibility of a CLD while incorporating accumulation, a key source of dynamic complexity [50], omitted in MARVEL and FCMs. The leverage point ranking under uncertainty and sensitivity analysis offers dynamic insight without the elicitation burden of MARVEL and FCM, where the properties of every connection must be specified. Delays are still an important source of dynamics, and in D2D, they arise only from accumulation. Specifying additional delays and relation strengths, as in MARVEL, could enrich D2D at the cost of extra model development time.

*3.6 Limitations of D2D*

D2D assumes that the CLD adequately captures the relevant system structure. In reality, important variables, connections, or feedback mechanisms may be inadvertently omitted—a limitation shared with system dynamics approaches in general, including those calibrated with empirical data. D2D further assumes linearity in causal relationships, which may not be appropriate for systems where nonlinearities, such as threshold effects, are central to behavior. Although D2D accommodates interaction terms (e.g., second-order interactions or quadratic effects), their appropriate inclusion requires both domain and modeling expertise, potentially compromising the method's simplicity and accessibility. A good use of D2D depends on CLD quality and accurate variable labeling, and misspecification can affect simulation results. Therefore, we recommend that users assess the sensitivity of results to structural assumptions, e.g., by re-running the analysis with plausible alternative CLD configurations or variable labels. D2D also assumes that system leverage points can be meaningfully assessed through one or more VOI(s) explicitly represented in the CLD. While D2D allows for multiple VOIs, it may not be suited to contexts where interventions target feedback loops, influence latent constructs, or alter system structure that is not easily mapped to specific variables. In this regard, sensitivity analysis on model parameters can also be interpreted as a simulated intervention on connections; however, this was not our focus here. Finally, we deliberately limited D2D functionality. Although D2D can be extended with more advanced analyses, such as stability analysis via eigenvalue decomposition or feedback loop analysis, we prioritized simulated interventions and sensitivity analysis to maintain conceptual simplicity, enhance accessibility, and support broader adoption.

Considering these limitations, D2D is best seen as a tool for exploring the implications of a CLD's structure in the face of uncertainty, not a substitute for fully specified and calibrated system dynamics modeling. It does not bypass the need for empirical estimation or detailed specification of functional forms. Rather, it provides a low-barrier entry point for early-stage quantification, exploration of interventions, and iterative model refinement. To encourage further testing and broaden access, the method is implemented in both an open-source Python package and a web-based application. While the initial results are promising, additional validation across diverse cases will be essential to establish D2D's utility and further develop its capabilities.

## 4. Conclusion

This paper introduced the D2D approach for generating quantitative insights from CLDs, even without empirical data. D2D enables simulations of hypothetical interventions under uncertainty and offers sensitivity analyses to support early-stage identification of leverage points and help prioritize data collection. Beyond exploration, D2D also serves as a foundation for the gradual development of fully specified SDMs. It accommodates more detailed knowledge of the system—such as interaction terms—and aligns naturally with Bayesian updating as data become available. In this way, D2D bridges the gap between conceptual mapping and advanced system dynamics modeling, offering researchers a low-barrier tool to deepen their understanding of complex systems.

## Abbreviations

D2D: Diagrams-to-Dynamics
CLD: Causal loop diagram
SDM: System dynamics model
VOI: Variable of interest
SD: Standard deviation

## Declarations

*Acknowledgments*

VVV and JFU acknowledge fruitful discussions with Wenying Liao and Christopher de Montgomery on the practical use of D2D.

*Consent for publication*

Not applicable.

*Funding*

JFU was supported by a Rubicon grant (04520232320007) of the Netherlands Organisation for Health Research and Development (ZonMw). The project was additionally supported by funding from the Lundbeck Foundation (grant no. R396-2022-352) and Trygfonden (grant no. 157232). The Copenhagen Health Complexity Center is funded by TrygFonden. NHR received funding from the European Union (ERC, LAYERS, project no. 101124807). Views and opinions expressed are, however, those of the author(s) only and do not necessarily reflect those of the European Union or the European Research Council. Neither the European Union nor the granting authority can be held responsible for them.

*Data availability*

Data sharing is not applicable to this article as no datasets were generated or analyzed during the current study. A link to the reproducible code for the example in this study is provided here: https://doi.org/10.17605/OSF.IO/4UTGW.

*Author contributions*

J.F.U, L.C. V.V.V, L.K.E. conceptualized the study. J.F.U conducted the analyses, prepared the figures, implemented the code, and wrote the first and subsequent drafts of the manuscript. V.V.V. and R.Q. advised on the development of the tool. L.C. V.V.V, L.K.E, N.H.R., G.A.V. critically reviewed and revised the manuscript. All authors read and approved the final manuscript.

*Competing interests*

The authors declare that they have no competing interests.

*Ethics approval and consent to participate*

Not applicable.